\documentclass[letterpaper, 10 pt, conference]{ieeeconf}
\usepackage{times}
\usepackage[pdftex]{graphicx}
\usepackage{subfigure}
\usepackage{amsmath,amssymb,amsopn,amstext,amsfonts}
\usepackage{cancel}
\usepackage[space]{cite}
\usepackage{pdfsync}
\usepackage{balance}
\usepackage{color}
\usepackage{mathtools}
\usepackage{makecell}
\usepackage{algorithm}
\usepackage{algorithmic}
\usepackage{bm}

\usepackage{diagbox}
\usepackage{float}
\usepackage{epstopdf}
\usepackage{pifont}
\usepackage{multirow}
\usepackage{url}
\usepackage{tabularx}
\usepackage[linkcolor=black,citecolor=black,urlcolor=black,colorlinks=true]{hyperref}

\usepackage{verbatim} 

\graphicspath{{../figure/}}
\DeclareGraphicsExtensions{.png,.jpg,.eps,.pdf}
\IEEEoverridecommandlockouts
\title{\LARGE \bf Distributed Swarm Trajectory Optimization for Formation Flight in Dense Environments }
\author{Lun Quan$^*$, Longji Yin$^*$, Chao Xu, and Fei Gao$^\dagger$
	\thanks{ $^*$Indicates equal contribution. }
	\thanks{$^\dagger$Corresponding author: {\tt\small fgaoaa@zju.edu.cn}.}
	\thanks{
	This work was supported by the National Natural Science Foundation of China under Grants 62003299. 
	All authors are from the State Key Laboratory of Industrial Control Technology, Institute of Cyber-Systems and Control, Zhejiang University, Hangzhou 310027, China, and the Huzhou Institute, Zhejiang University, Huzhou 313000, China.
	}
}
\begin{document}

\maketitle
\thispagestyle{empty}
\pagestyle{empty}

\begin{abstract}
For aerial swarms, navigation in a prescribed formation is widely practiced in various scenarios. However, the associated planning strategies typically lack the capability of avoiding obstacles in cluttered environments. To address this deficiency, we present an optimization-based method that ensures collision-free trajectory generation for formation flight.
In this paper, a novel differentiable metric is proposed to quantify the overall similarity distance between formations. We then formulate this metric into an optimization framework, which achieves spatial-temporal planning using polynomial trajectories. Minimization over collision penalty is also incorporated into the framework, so that formation preservation and obstacle avoidance can be handled simultaneously. To validate the efficiency of our method, we conduct benchmark comparisons with other cutting-edge works. Integrated with an autonomous distributed aerial swarm system, the proposed method demonstrates its efficiency and robustness in real-world experiments with obstacle-rich surroundings\footnote{https://www.youtube.com/watch?v=lFumt0rJci4}. We will release the source code for the reference of the community\footnote{https://github.com/ZJU-FAST-Lab/Swarm-Formation\label{git}}.
\end{abstract}

\section{Introduction}
\label{sec:introduction}
Autonomous aerial swarms can be employed for many systematic and cooperative tasks, such as search and rescue\cite{marconi2012sherpa}, collaborative mapping\cite{mahdoui2020communicating}, and package delivery\cite{dorling2016vehicle}. In some scenarios, it could be desired that the swarm moves according to a specified formation. For example, in \cite{jahn2017distributed}, robots are required to form and maintain a virtual fence for animal herding tasks.

Extensive research works exist for aerial swarm navigation in formation, but none of them achieves robust formation flight in obstacle-dense environments. In practice, robots are repulsed to deviate from obstacles for safety, while formation imposes tracking targets that may oppose the obstacle avoidance. How to systematically trade off these two conflicting requirements is the key point to accomplish non-colliding formation flights. In the literature\cite{oh2015survey}, consensus-based local control laws are widely used for formation maneuvering. However, the local control scheme is inherently incapable of planning over a prediction horizon, which severely undermines its practicality in complex environments. A group of optimization-based works maintain the formation by enforcing relative position constraints on each agent. Nevertheless, with this strategy, the formation would passively yield to the obstacles in dense scenarios since the surroundings could always defy the positional constraints. To summarize, a swarm trajectory planning method that can effectively manage both formation and obstacle avoidance in dense environments is lacking in the literature.

To bridge the gap, we propose a swarm trajectory optimization method capable of navigating swarms in formation while avoiding obstacles. We model the formation using undirected graphs and define a differentiable Laplacian-based metric that assesses the difference between formation shapes in three-dimensional workspaces. Rather than independently inspecting each agent's tracking error, our metric quantitatively evaluates the overall performance of formation maintenance, and provides greater flexibility for formation maneuvering by virtue of its invariance to translation, rotation, and scaling. To formulate the trade-offs between formation and obstacle avoidance, we design an unconstrained optimization framework that simultaneously optimizes the trajectories over feasibility cost, collision penalty, and formation similarity error. Benchmark comparisons are carried out with other state-of-the-art methods. Finally, to verify that our method is efficient and practical, extensive experiments are conducted on a real distributed aerial swarm system integrated with the proposed method. 

\begin{figure}[t]
    \begin{center}
        \includegraphics[width=1.0\columnwidth]{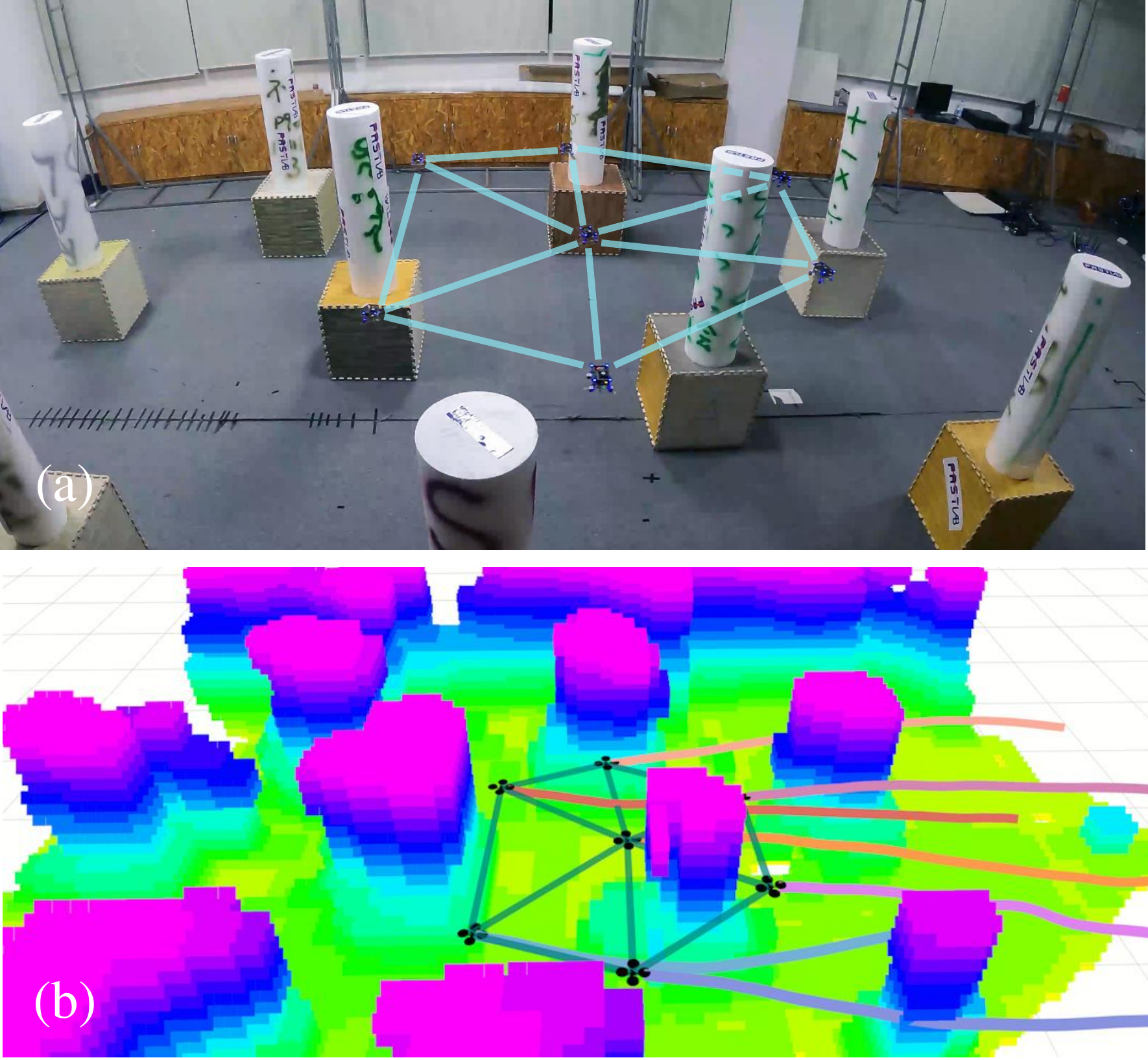}
    \end{center}
    \caption{A swarm of seven quadrotors in a regular hexagon formation is traversing an unknown obstacle-rich area. (a)\,\,: A snapshot of the formation flight. (b)\,\,: The visualization of the local map and executed trajectories. It should be pointed out that in this paper, the blue line segments only represent the outlines of the shape, but not the connectivity of the graph.}
    \label{fig:main}
    \vspace{-1.0cm}
\end{figure}

We summarize our contributions as: 
\begin{enumerate}
	\item A differentiable graph-theory-based cost function that quantifies the similarity distance between three-dimensional formations. 
	\item A distributed optimization framework with a joint cost function that takes formation similarity, obstacle avoidance, and dynamic feasibility into account, which makes the swarm has the ability to balance moving in formation while avoiding obstacles.
	\item A series of simulations and real-world experiments with a distributed aerial swarm system that validates the efficiency and robustness of our method. The source code is released for the reference of the community\textsuperscript{\ref {git}}.
\end{enumerate}

\section{Related works}
\label{sec:related_works}

\subsection{Distributed Swarm Trajectory Planning}
\label{Multicopter swarm}
Extensive works exist for trajectory planning of distributed swarms. 
The concept of VO (velocity obstacle) is leveraged and generalized by Van Den Berg et al.\cite{van2011RR,van2011reciprocal,bareiss2013reciprocal} to accomplish reciprocal collision avoidance for multiple robots. 
However, the smoothness of the resulting trajectories cannot be guaranteed by VO-based approaches, which greatly impairs the usability on real robot systems.

In order to produce high-quality collision-free trajectories, optimization-based methods are widely introduced in the literature on distributed multicopter swarms\cite{arul2020dcad,luis2019trajectory,park2020efficient}. Zhou et al.\cite{zhou2017fast} incorporate Voronoi cell tessellation into a receding horizon QP scheme to prevent collision among the robots while planning. 
In \cite{chen2015decoupled}, Chen et al. employ SCP to address the multi-agent planning problem in non-convex space by incrementally tightening the collision constraints. Baca et al.\cite{baca2018model} combine MPC with a conflict resolution strategy to ensure mutual collision avoidance for outdoor swarm operations. Nevertheless, the computational load of the above optimation-based methods is large, which could hamper the applicability of the planners in highly dense scenarios.   

Recently, Zhou et al.\cite{zhou2021decentralized} present a distributed autonomous quadrotor swarm system using spatial-temporal trajectory optimization, which generates collision-free motions in dense environments merely in milliseconds. Our swarm trajectory generation scheme is based on this work.

\subsection{Multi-robot Navigation in Formation}
\label{Formation Maintenance}
Various techniques have been proposed to achieve multi-robot navigation in formation, which include virtual structures\cite{lewis1997high}, navigation functions\cite{de2006formation}, reactive behaviors\cite{balch1998behavior}, and consensus-based local control laws\cite{lin2013leader}. However, most of the existing methods only consider obstacle-free cases.

A group of works handle the formation navigation in constrained scenarios by designing feedback laws. 
Han et al.\cite{han2013local} present a formation controller based on complex-valued graph Laplacians. The formation scale is regulated by a leader to perform intended swarm maneuvering, like passing a corridor. In\cite{zhao2018affine}, Zhao proposes a leader-follower control law with which the formation can be affinely transformed in responding to the environmental changes. Most control approaches rely on the leader-follower scheme, where the formation parameters are only accessible to the leader.

Compared to the leader-follower scheme in\cite{han2013local,zhao2018affine}, fully decentralized strategies possess better scalability and resiliency to partial failures. 
In\cite{alonso2016distributed}, Alonso-Mora et al. control a formation of drones to avoid collision by optimally rearranging the desired formation and then planning local trajectories.
But since no inter-vehicle coordination exists in the distributed planners, formation maintenance is not conducted in the local planning phase. 
Zhou et al.\cite{zhou2018agile} combine virtual structure with potential fields to produce non-colliding trajectories for formation flight. Nevertheless, planning with multiple interacting vector fields is prone to deadlocks. Besides, trajectory optimality is neglected by their method.
Parys et al.\cite{van2017distributed} use DMPC to tackle the formation preservation by imposing relative position constraints on the swarm. In their framework, coordination among the agents passively breaks once the positional constraints are violated by the obstacles. In contrast, we formulate the overall formation requirement with a differentiable metric, which enables our optimizer to trade off formation and obstacle avoidance collectively and simultaneously.

\section{A Differentiable Formation similarity metric}
\label{sec:formation similarity}
A formation of $N$ robots is modeled by an undirected graph $\mathcal{G} = (\mathcal{V,E})$, where $\mathcal{V}:=\{1,2,...,N\}$ is the set of vertices, and $\mathcal{E} \subset \mathcal{V} \times \mathcal{V}$ is the set of edges. In graph $\mathcal{G}$, the vertex $i$ represents the $i^{th}$ robot with position vector $\mathbf{p}_i = [x_i,y_i,z_i] \in \mathbb{R}^3$ . An edge $e_{ij} \in \mathcal{E}$ that connects vertex $i\in \mathcal{V}$ and vertex $j\in \mathcal{V}$ means the robot $i$ and $j$ can measure the geometric distance between each other. In our work, each robot communicates with all other robots, thus the formation graph $\mathcal{G}$ is complete. Each edge of the graph $\mathcal{G}$ is associated with a non-negative number as a weight. In this work, the weight of edge $e_{ij}$ is given by
\begin{equation}
\label{graph_weight}
    w_{ij} =  \parallel\mathbf{p}_i-\mathbf{p}_j\parallel^2,\;\; (i,j) \in \mathcal{E},
\end{equation}
where $\parallel\cdot\parallel$ denotes the Euclidean norm.  
Now the adjacency matrix $\mathbf{A} \in \mathbb{R}^{N\times N}$ and degree matrix $\mathbf{D}\in \mathbb{R}^{N\times N}$ of the formation $\mathcal{G}$ is determined. Thus, the corresponding Laplacian matrix is given by 
\begin{equation}
 \mathbf{L} = \mathbf{D} - \mathbf{A}.
\end{equation}
With the above matrices, the symmetric normalized laplacian matrix of graph $\mathcal{G}$ is defined as
\begin{equation}
\label{L_normalization}
    \mathbf{\hat{L}} = \mathbf{D}^{-1/2}\mathbf{L}\mathbf{D}^{-1/2} = \mathbf{I} - \mathbf{D}^{-1/2}\mathbf{A}\mathbf{D}^{-1/2},
\end{equation}
where $\mathbf{I} \in \mathbb{R}^{N\times N}$ is the identity matrix. 

As a graph representation matrix, Laplacian contains information about the graph structure\cite{NatureReport}. To achieve the desired swarm formation, we propose a formation similarity distance metric as
\begin{equation}
\label{L_metric}
    f = \parallel\mathbf{\hat{L}}-\mathbf{\hat{L}}_{des}\parallel^2_F = tr\{(\mathbf{\hat{L}}-\mathbf{\hat{L}}_{des})^T(\mathbf{\hat{L}}-\mathbf{\hat{L}}_{des})\},
\end{equation}
where $tr\{\cdot\}$ denotes the trace of a matrix, $\mathbf{\hat{L}}$ is the symmetric normalized Laplacian of the current swarm formation, $\mathbf{\hat{L}}_{des}$ is the counterpart of the desired formation. Frobenius norm $\parallel\cdot\parallel_F$ is used in our distance metric. $f$ is natively invariant to translation and rotation of the formation, since the corresponding graph is weighted by the absolute distance between robot positions. Scaling invariance is achieved by normalizing graph Laplacian with the degree matrix in (\ref{L_normalization}).
\begin{figure}
    \begin{center}
         \includegraphics[width=1.0\columnwidth]{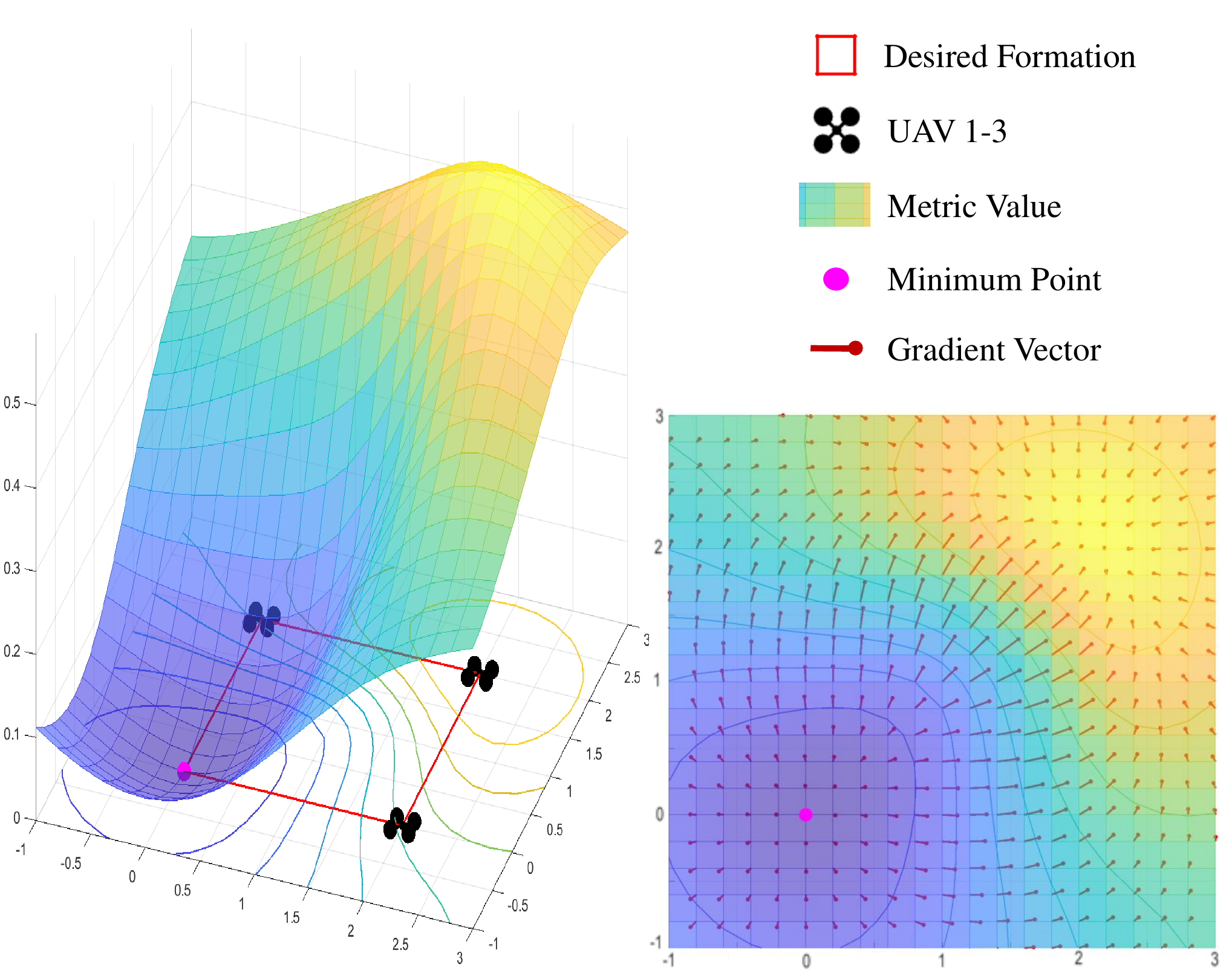}
    \end{center}
    \caption{Illustration of the formation similarity metric and its gradients. A square formation is desired by the swarm. The surface shows the profile of similarity metric when one UAV moves in the plane and the other three remain still. The minimum suggests the best position for the UAV to form the desired shape. Right side shows the gradients of the similarity metric.}
    \label{fig:metric_gradient}
    \vspace{-0.7cm}
\end{figure}

Our metric is analytically differentiable with respect to the position of each robot. For robot $i$, we use the weights of its $n$ adjacent edges $\{e_{i1},\,e_{i2},\,...,\;,e_{in}\}$ to form a weight vector $\mathbf{w}_{i} = [w_{i1},\,w_{i2},\,...,\;,w_{in}]^T$. By chain rule, the gradient of $f$ with respect to $\mathbf{p}_i$ can be written as 
\begin{equation}
    \label{eq:f_p}
    \frac{\partial f}{\partial \mathbf{p}_i} = \frac{\partial \mathbf{w}_i}{\partial \mathbf{p}_i} \frac{\partial f}{\partial \mathbf{w}_i}.
\end{equation}
According to our metric (\ref{L_metric}), the gradient of $f$ with respect to each weight $w_{ij}$ can be computed as follow
\begin{equation}
    \frac{\partial f}{\partial w_{ij}} = tr\{(\frac{\partial f}{\partial \hat{\mathbf{L}}})^T(\frac{\partial \hat{\mathbf{L}}}{\partial w_{ij}})\},
\end{equation}
where
\begin{equation}
\frac{\partial f}{\partial \mathbf{\hat{L}}} = \frac{\partial ||\mathbf{\hat{L}}-\mathbf{\hat{L}}_{des}||_F^2}{\partial \mathbf{\hat{L}}} =
2(\mathbf{\hat{L}}-\mathbf{\hat{L}}_{des}),
\end{equation}
\begin{equation}
    \frac{\partial \mathbf{\hat{L}}}{\partial w_{ij}} =
    -\frac{\partial(\mathbf{D}^{-1/2}\mathbf{A}\mathbf{D}^{-1/2})}{\partial w_{ij}}.
\end{equation}
Then the gradient $\partial f/\partial \mathbf{w}_i$ can be written as 
\begin{equation}
   \partial f/\partial \mathbf{w}_i = [\partial f/\partial w_{i1}, \partial f/\partial w_{i2}, ..., \partial f/\partial w_{in}]^T.
\end{equation}
As for $\partial \mathbf{w}_i/\partial \mathbf{p}_i$, the Jacobian can be easily derived since the weight function (\ref{graph_weight}) is differentiable. 
Fig.\ref{fig:metric_gradient} shows a profile of the metric and the gradient for a square formation. 
\section{Spatial-temporal trajectory optimization for formation flight}
\label{sec:optimization problem}

\subsection{Trajectory Representation}
\label{subsec:representation}
In this work, we adopt the MINCO representation~\cite{wang2021geometrically}, a minimum control effort polynomial trajectory class to conduct spatial-temporal deformation of the flat-output trajectory

\begin{equation}
	\label{equ:minco}
	\begin{aligned}
		\Xi_{MINCO}= & \{p(t):[0,T_{\Sigma}]\mapsto\mathbb{R}^m | \textbf{c}=\textsl{C}(\textbf{q},\textbf{T}), \\
		& \textbf{q}\in\mathbb{R}^{m(M-1)},\textbf{T}\in\mathbb{R}_{>0}^M\},
	\end{aligned}
\end{equation}
where $\textbf{c}=(c_1^T,\cdots,c_M^T)^T$ is the polynomial coefficient, $\textbf{q}=(q_1,\cdots,q_{M-1})$ the intermediate points, $\textbf{T}=(T_1,\cdots,T_M)^T$ the time vector, $\textsl{C}(\textbf{q},\textbf{T})$ the parameter mapping constructed from Theorem 2 in~\cite{wang2021geometrically}, and $T_{\Sigma}=\sum_{i=1}^MT_i$ the total time.

A $m$-dimensional $M$-piece trajectory $p(t)$ is defined as
\begin{equation}
	\label{equ:M-piece trajectory}
	p(t)=p_i(t-t_{i-1}),~~\forall{t}\in[t_{i-1},t_i),
\end{equation}
and the $i^{th}$ piece trajectory is represented by a $N=5$ degree polynomial
\begin{equation}
	\label{equ:i-th piece}
	p_i(t)=c_i^T\beta(t),~~\forall{t}\in[0,T_i],
\end{equation}
where $c_i\in \mathbb{R}^{6\times m}$ is the coefficient matrix, $\beta(t)=[1,t,\cdots,t^N]^T$ is the natural basis, and $T_i=t_i-t_{i-1}$ is the time allocation for the $i^{th}$ piece.

MINCO is uniquely determined by $(\textbf{q},\textbf{T})$. And the parameter mapping $\textbf{c}=\textsl{C}(\textbf{q},\textbf{T})$ converts trajectory representations $(\textbf{c},\textbf{T})$ to $(\textbf{q},\textbf{T})$ with linear time and space complexity, which allows any second-order continuous cost function $J(\textbf{c},\textbf{T})$ to be represented by $\widetilde{J}(\textbf{q},\textbf{T})$. 
Hence, $\partial\widetilde{J}/\partial\textbf{q}$ and $\partial\widetilde{J}/\partial\textbf{T}$ can be efficiently obtained from $\partial{J}/\partial\textbf{c}$ and $\partial{J}/\partial\textbf{T}$, respectively.

Especially, in order to handle the time integral constraints $\psi(p(t),\cdots,p^{(3)}(t))\preceq\textbf{0}$, such as collision avoidance and dynamical feasibility, we transform them into finite-dimensional constraints $\psi(\hat{p}_{i,j})$ by sampling \textbf{constraint points} $\hat{p}_{i,j}=p_i((j/\kappa_i)\cdot T_i)$ on the trajectory, where $\kappa_i$ is the sample number on the $i^{th}$ piece.

\subsection{Optimization Problem Formulation}
\label{subsec:problem}
We formulate the trajectory generation for formation flight as an unconstrained optimization problem
\begin{equation}
	\label{equ:minJ}
	\min_{\textbf{c},\textbf{T}}\;[J_e,J_t,J_o,J_f,J_r,J_d,J_u]\cdot\lambda,
\end{equation}
where $\lambda$ is the weight vector to trade off each cost function.

\subsubsection{Control Effort $J_e$}
\label{sub:Je}
The third order control input for the $i^{th}$ piece trajectory and its gradients are written as 
\begin{equation}
	\label{equ:Je}
	J_e=\int_0^{T_i}\parallel p_i^{(3)}(t)\parallel^2dt,
\end{equation}
\begin{equation}
	\label{equ:Je_c}
	\frac{\partial J_e}{\partial c_i}=2\left( \int_0^{T_i}\beta^{(3)}(t)\beta^{(3)}(t)^Tdt \right)c_i,
\end{equation}
\begin{equation}
	\label{equ:Je_T}
	\frac{\partial J_e}{\partial T_i}=c_i^T\beta^{(3)}(T_i)\beta^{(3)}(T_i)^Tc_i.
\end{equation}

\subsubsection{Total Time $J_t$}
\label{sub:Jt}
In order to ensure the aggressiveness of the trajectory, we minimize the total time $J_t=\sum_{i=1}^MT_i$. The gradients are given by $\partial J_t/\partial \textbf{c}=\textbf{0}$ and $\partial J_t/\partial \textbf{T}=\textbf{1}$.

\subsubsection{Obstacle Avoidance $J_o$}
\label{sub:Jo}
Inspired by\cite{zhou2019robust}, obstacle avoidance penalty $J_o$ is computed using Euclidean Signed Distance Field (ESDF). The constraint points which are close to the obstacles are selected by
\begin{equation}
	\label{equ:psi_o}
	\psi_o(\hat{p}_{i,j})=
	\begin{cases}
		d_{thr} - d(\hat{p}_{i,j}), & \mbox{if} \quad d(\hat{p}_{i,j})<d_{thr}, \\
		0, & \mbox{if} \quad d(\hat{p}_{i,j}) \ge d_{thr},
	\end{cases}
\end{equation}
where $d_{thr}$ is the safety threshold and $d(\hat{p}_{i,j})$ is the distance between the considered point and the closest obstacle around it.
Then the obstacle avoidance penalty is obtained by computing the weighted sum of sampled constraint function:
\begin{equation}
	\label{equ:Jo}
	J_o = \frac{T_i}{\kappa_i} \sum_{j=0}^{\kappa_i} \bar{\omega}_j \max\{\psi_o(\hat{p}_{i,j}),0\}^3,
\end{equation}
where $(\bar{\omega}_0,\bar{\omega}_1,\cdots,\bar{\omega}_{\kappa_i-1},\bar{\omega}_{\kappa_i})=(1/2,1,\cdots,1,1/2)$ are the orthogonal coefficients following the trapezoidal rule~\cite{Press2007numerical}.

The gradients of $J_o$ w.r.t $c_i$ and $T_i$ are detailed as 
\begin{equation}
	\label{equ:dJo_dci}
	\frac{\partial J_o}{\partial c_i} = \frac{\partial J_o}{\partial \psi_o} \frac{\partial \psi_o}{\partial c_i},
\end{equation}
\begin{equation}
	\label{equ:dJo_dTi}
	\frac{\partial J_o}{\partial T_i} = \frac{J_o}{T_i} + \frac{\partial J_o}{\partial \psi_o} \frac{\partial \psi_o}{\partial t} \frac{\partial t}{\partial T_i},
\end{equation}
\begin{equation}
	\label{equ:dt_dTi}
	\frac{\partial t}{\partial T_i}=\frac{j}{\kappa_i}, \qquad t=\frac{j}{\kappa_i}T_i,
\end{equation}
where $t$ is the relative time on the piece. For the case that $d(\hat{p}_{i,j})<d_{thr}$ , the gradients are given by
\begin{equation}
	\label{psi_o_gradient}
	\frac{\partial \psi_o}{\partial c_i}=-\beta(t) \nabla d^T, \qquad 
	\frac{\partial \psi_o}{\partial t}=-\nabla d^T\dot{p}(t),
\end{equation}
where $\nabla d$ is the gradient of ESDF in $\hat{p}_{i,j}$. Otherwise, the gradients become $\partial \psi_o/\partial c_i=\textbf{0},~\partial \psi_o/\partial t=0$.

\subsubsection{Swarm Formation Similarity $J_f$}
\label{sub:Jf}
In Sec.\ref{sec:formation similarity}, we design a differentiable metric to quantify the similarity distance between swarm formations. In optimization, the similarity error between the current formation and the desired formation is measured by $\psi_f = f(p(t),\bigcup_{\Phi}p_\phi(\tau))$, where $f(\cdot)$ is detailed in (\ref{L_metric}) and $\Phi$ represents the collection of other agents. 

Since $J_f$ involves the trajectories of other agents, we need to deal with both the relative time $t=j T_i/ \kappa_i$ of the own trajectory and the global timestamp $\tau=T_1+...+T_{i-1}+j T_i/ \kappa_i$ of others' trajectories. $J_f$ considers the preceding time $T_l$ for any $1\leq l \leq i$ and is formulated as
\begin{equation}
	\label{eq:J_f}
	J_f = \frac{T_i}{\kappa_i} \sum_{j=0}^{\kappa_i} \bar{\omega}_j \max\{\psi_f(p(t),\bigcup_{\Phi}p_\phi(\tau)),0\}^3.
\end{equation}
The gradients of $J_f$ w.r.t $c_i$ and $T_l$ are computed as 
\begin{equation}
	\label{eq:dJf_dci}
	\frac{\partial J_f}{\partial c_i} = \frac{\partial J_f}{\partial \psi_f} \frac{\partial \psi_f}{\partial c_i},
\end{equation}
\begin{equation}
	\label{eq:dJf_dTl} 
	\frac{\partial J_f}{\partial T_l} = \frac{J_f}{T_l}+\frac{\partial J_f}{\partial \psi_f}\frac{\partial \psi_f}{\partial T_l}. 
\end{equation}
To derive $\partial \psi_f/\partial T_l$, $\psi_f$ need to be differentiated by $t$ and $\tau$:
\begin{equation}
	\label{eq:dpsi_f_dTl}
	\frac{\partial \psi_f}{\partial T_l} = \frac{\partial \psi_f}{\partial t}\frac{\partial t}{\partial T_l} + \frac{\partial \psi_f}{\partial \tau}\frac{\partial \tau}{\partial T_l},
\end{equation}
\begin{equation}
	\label{eq:dt_dTl}
	\frac{\partial t}{\partial T_l} = 
	\begin{cases}
		\frac{j}{\kappa_i}, &l=i, \\
		0                 , &l<i,
	\end{cases} \qquad
	\frac{\partial \tau}{\partial T_l} =
	\begin{cases}
		\frac{j}{\kappa_i}, &l=i, \\
		1                 , &l<i.
	\end{cases}
\end{equation}
The gradients of $\psi_f$ w.r.t $c_i$, $t$ and $\tau$ are given by
\begin{equation}
	\label{eq:dgf_dc}
	\frac{\partial \psi_f}{\partial c_i} =\frac{\partial \psi_f}{\partial p(t)} \frac{\partial p(t)}{\partial c_i},
\end{equation}
\begin{equation}
	\label{eq:dgf_dt}
	\frac{\partial \psi_f}{\partial t} = \frac{\partial \psi_f}{\partial p(t)} \frac{\partial p(t)}{\partial t} = \frac{\partial \psi_f}{\partial p(t)} \dot{p}(t),
\end{equation}
\begin{equation}
	\label{eq:dgf_dtau}
	\frac{\partial \psi_f}{\partial \tau} = \sum_{\Phi} \frac{\partial \psi_f}{\partial p_\phi(\tau)} \frac{\partial p_\phi(\tau)}{\partial \tau} = \sum_{\Phi}\frac{\partial \psi_f}{\partial p_\phi(\tau)} \dot{p}_\phi(\tau),
\end{equation}
where the gradient of $\psi_f$ to $p(t)$ and $p_\phi(\tau)$ is detailed in (\ref{eq:f_p}).

\subsubsection{Swarm Reciprocal Avoidance $J_r$}
\label{sub:Jr}
We penalize the constraint points which are close to other agents' trajectories at global timestamp $\tau$. Thus, the cost function of swarm reciprocal avoidance is defined as
\begin{equation}
	\label{eq:Jr}
	J_r = \sum_{\Phi} \frac{T_i}{\kappa_i} \sum_{j=0}^{\kappa_i} \bar{\omega}_j \max\{\psi_{r_\phi}(p(t),\tau),0\}^3,
\end{equation}
\begin{equation}
	\label{eq:psi_phi}
	\psi_{r_\phi}(p(t),\tau)=D_r^2-d(p(t),p_\phi(\tau))^2,
\end{equation}
\begin{equation}
	\label{eq:phi_d}
	d(p(t),p_\phi(\tau))=\parallel p(t)-p_\phi(\tau)\parallel,
\end{equation}
where $D_r$ is the clearance between each agent.

The gradients of $J_r$ w.r.t $c_i$ and $T_l$ are the same as (\ref{eq:dJf_dci}) and (\ref{eq:dJf_dTl}), and $\partial \psi_{r_\phi}/\partial T_l$ is the same as (\ref{eq:dpsi_f_dTl}). 
When $D_r^2 \geq d(p(t),p_\phi(\tau))^2$, the gradients of $\psi_{r_\phi}$ w.r.t $c_i$, $t$ and $\tau$ are
\begin{equation}
	\label{eq:phi_r_ci}
	\frac{\partial \psi_{r_\phi}}{\partial c_i}=-2\beta(t)(p(t)-p_{\phi}(\tau))^T,
\end{equation}
\begin{equation}
	\label{eq:phi_r_t}
	\frac{\partial \psi_{r_\phi}}{\partial t}=-2(p(t)-p_{\phi}(\tau))^T\dot{p}(t),
\end{equation}
\begin{equation}
	\label{eq:phi_r_tau}
	\frac{\partial \psi_{r_\phi}}{\partial \tau}=2(p(t)-p_{\phi}(\tau))^T\dot{p_{\phi}}(t).
\end{equation}

\subsubsection{Dynamical Feasibility $J_d$}
\label{sub:Jd}
We limit the maximum value of velocity, acceleration, and jerk to guarantee that the trajectory can be executed by the agent.
Readers can refer to \cite{zhou2021decentralized} for more details.

\subsubsection{Uniform Distribution of Constraint Points $J_u$}
\label{sub:Ju}
The constraint points are expected to be space-uniform. Non-uniform constraint points may skip some small-sized obstacles, which could diminish the safety of the resulting trajectory.
Therefore, the uniform distribution penalty $J_u$ is optimized to prevent constraint points from gathering in certain locations. Readers can refer to \cite{zhou2021decentralized} for more details.

\begin{figure}
    \begin{center}
        \includegraphics[width=1.0\columnwidth]{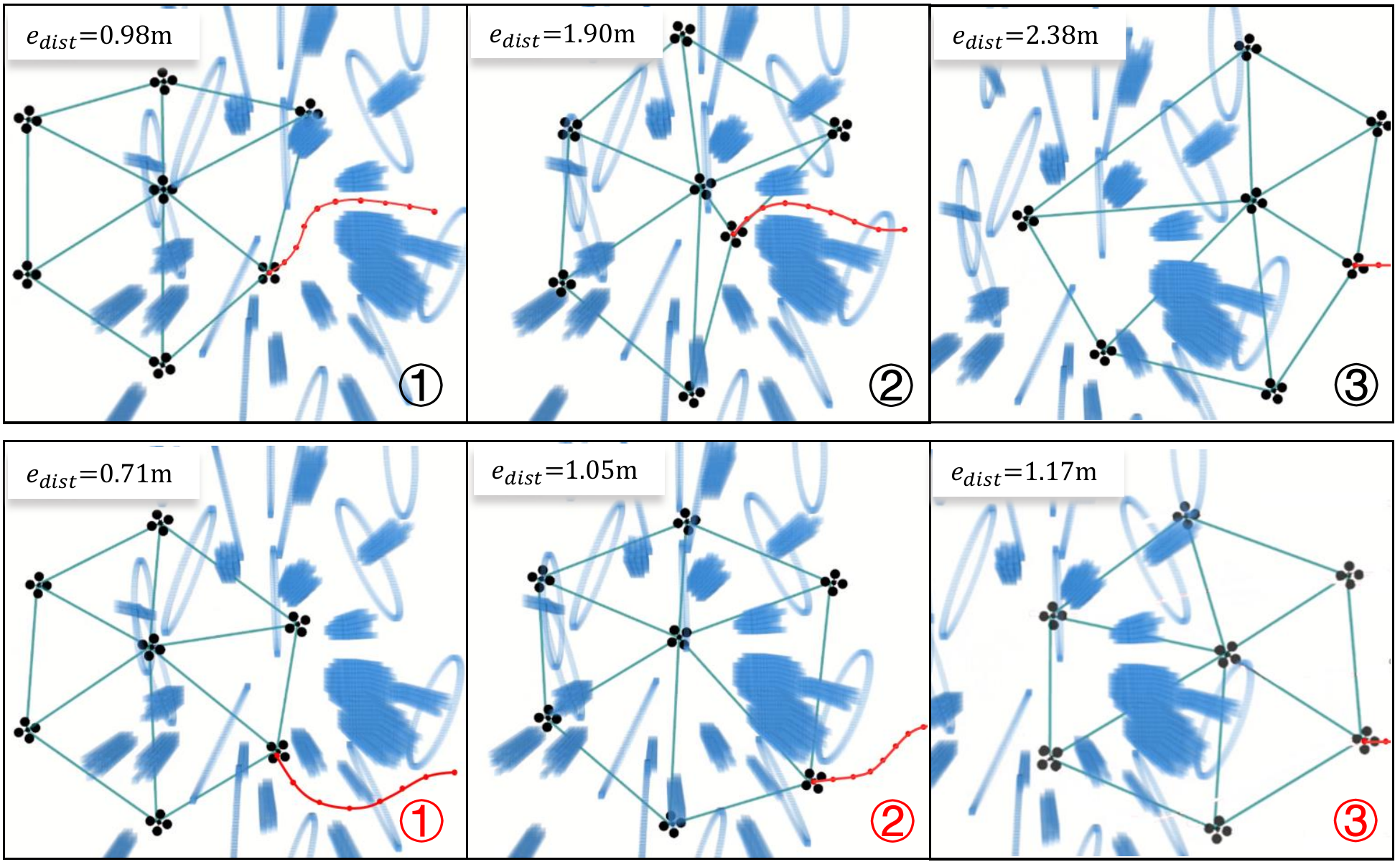}
    \end{center}
    \caption{ Benchmark comparison between Turpin's method (the upper row) and our method (the lower row). With Turpin's method, robots could take non-colliding routes that actually impairs the formation. In contrast, with our method, robots make more reasonable decisions to hold the formation. Position error $e_{dist}$ of each formation is shown in the figure.}
    \label{fig:benchmark}
\end{figure}

\begin{table}
\caption{Formation Navigation Methods Comparison}
\label{benchmark table}
\renewcommand\arraystretch{1.2}
\begin{tabular}{|c|c|c|c|c|}
\hline
     Scenario & Method & success rate(\%) & $e_{dist}$(m) & $e_{sim}$ \\ \hline
    & Zhou's\cite{zhou2018agile} & 65 & 1.47 & 0.0162\\ \cline{2-5} 
     Sparse & Turpin's\cite{turpin2012trajectory} & 85 & 1.08 &0.0066 \\ \cline{2-5}
   & \textbf{Ours} & \bf100 & \bf0.77 &\bf0.0032 \\ \hline
   & Zhou's\cite{zhou2018agile} & 15 & - &- \\ \cline{2-5} 
     Medium & Turpin's\cite{turpin2012trajectory} & 75 & 1.55 &0.0162 \\ \cline{2-5}
   & \textbf{Ours} & \bf100 & \bf0.90 &\bf0.0045 \\ \hline
   & Zhou's\cite{zhou2018agile} & 0 & - &- \\ \cline{2-5} 
     Dense& Turpin's\cite{turpin2012trajectory} & 60 & 2.01 &0.0278 \\ \cline{2-5}
   & \textbf{Ours} & \bf95 & \bf1.25 &\bf0.0107 \\ \hline
\end{tabular}
\end{table}

\section{Benchmark}
\label{sec:benchmark}
To demonstrate the efficiency and robustness of our method, benchmark comparisons are conducted with cutting-edge formation control methods. We compare our work with Zhou's method\cite{zhou2018agile} and Turpin's method\cite{turpin2012trajectory}. We implement the formation control method in Turpin's work and adapt it to the dense environments by adding our own obstacle avoidance strategy. However, unlike our work, in \cite{zhou2018agile} and \cite{turpin2012trajectory}, changing the scale and rotation of the formation is not permitted during the flight. Hence, to evaluate the performance fairly, a new indicator of overall position error is proposed for formation flight. 

Inspired by \cite{parker2018pipeline}, in order to assess the overall position error $e_{dist}$ between the current formation $\mathcal{F}^c$ and the desired one $\mathcal{F}^d$, we solve the following nonlinear optimization problem to find the best similarity transformation ($Sim(3)$ transformation) that aligns $\mathcal{F}^c$ with $\mathcal{F}^d$:
\begin{equation}
    \label{iso_optimization}
    e_{dist} = \displaystyle{\min_{\mathbf{R},\,\mathbf{t},\,s} \sum_{i=1}^n ||\textbf{p}_{i}^d -  (s\,\textbf{R}\,\textbf{p}^c_i + \textbf{t})||^2}.
\end{equation}
Notations $\textbf{p}_{i}^d$ and $\textbf{p}_{i}^c$ represent the position of $i^{th}$ robot in formation $\mathcal{F}^d$ and $\mathcal{F}^c$, respectively. The $Sim(3)$ transformation is composed of a rotation $\mathbf{R} \in SO(3)$, a translation $\mathbf{t}\in \mathbb{R}^3$ and a scale expansion $s \in \mathbb{R}_+$. By optimizing the transformation in (\ref{iso_optimization}) and applying to formations, the influence of scaling and rotation is squeezed out, so that all the formations can be equitably rated by measuring the position error w.r.t the desired formation. The larger the error $e_{dist}$, the more that $\mathcal{F}^c$ deviates from the desired shape $\mathcal{F}^d$. Besides the position error $e_{dist}$, methods are also compared over the success rate and the formation similarity error $e_{sim}$ we proposed in Sec.\ref{sec:formation similarity}.

We simulate seven drones flying in a regular hexagon formation from one side of an obstacle-rich map to another side with a velocity limit of 0.5$m/s$. The cluttered area is of 30$\times$15m size, and three obstacle densities are tested for comparison. Parameters are finely tuned for the best performance of each compared method.    

\begin{figure}[t]
    \begin{center}
         \includegraphics[width=1.0\columnwidth]{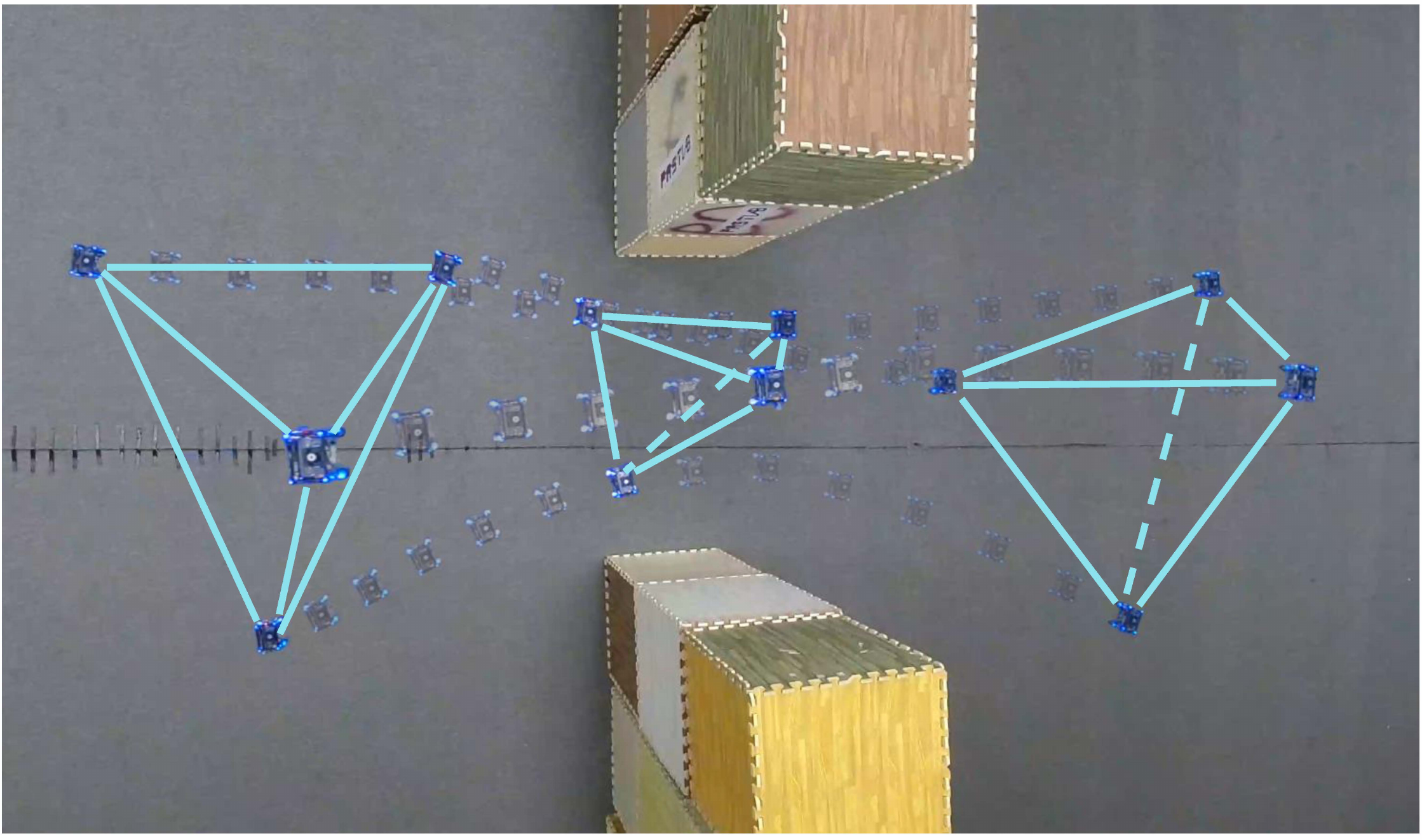}
    \end{center}
    \caption{Composite snapshots of a regular tetrahedron formation passing through a corridor. The swarm flies through the narrow space from right to left.}\label{fig:gap}
    \vspace{-1.5cm}
\end{figure}

The result is summarized in Table \ref{benchmark table}. It states that the success rate of Zhou's method\cite{zhou2018agile} is unsatisfactory in the presence of dense obstacles. The multiple interacting potential fields used in their work tend to generate local minima near the corridors. Thus, drones are often trapped in deadlocks. Turpin's method\cite{turpin2012trajectory} maintains formation by assigning desired relative positions to each pair of agents and then minimizing the relative error. However, these constraints are barely satisfied in cluttered scenarios, which raises the difficulty of finding feasible solutions and results in the lower success rates in Table \ref{benchmark table}. Furthermore, Turpin's strategy focuses on the individual tracking error instead of the overall formation shape, which could result in unfavorable trajectories for formation flights, as shown in Fig.\ref{fig:benchmark}. 

In Table \ref{benchmark table}, our method achieves better performance in terms of position error $e_{dist}$ and formation similarity error $e_{sim}$. Moreover, our success rate is promising even with complex surroundings, since the scaling and rotational invariance of the proposed metric provides more flexibility for the planner to generate motion plans.

\section{Real world and simulation experiments}
\label{sec:result}

\begin{figure*}[t]
    \begin{center}
        \includegraphics[width=2.0\columnwidth]{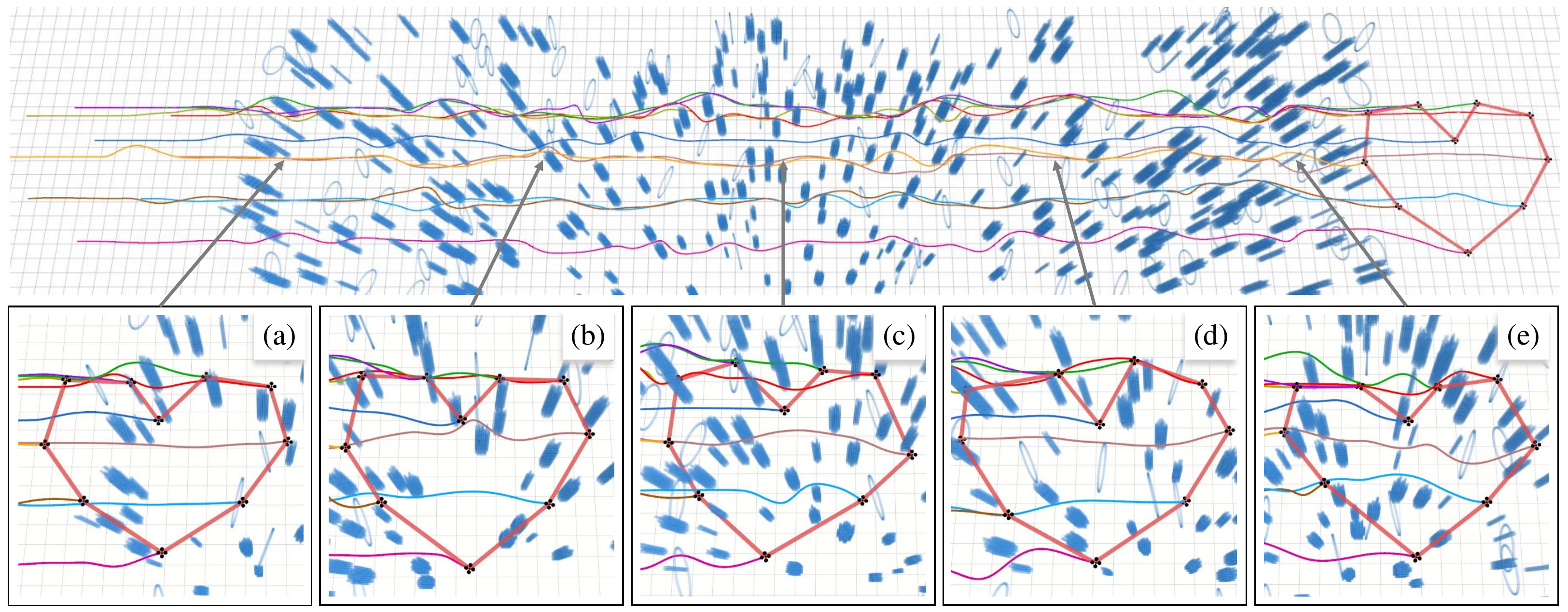}
    \end{center}
    \caption{A large-scale heart-shaped formation consisting of ten quadrotors traverses an unknown dense environment from the left side to right side.
    Snapshots (a)-(e) are uniformly sampled in the cluttered area.
    Colored curves show the executed trajectories of the formation flight.}
    \label{fig:heart}
\end{figure*}

\subsection{Implementation Details}
\label{sec:implementation}
\begin{figure}
    \begin{center}
        \includegraphics[width=1.0\columnwidth]{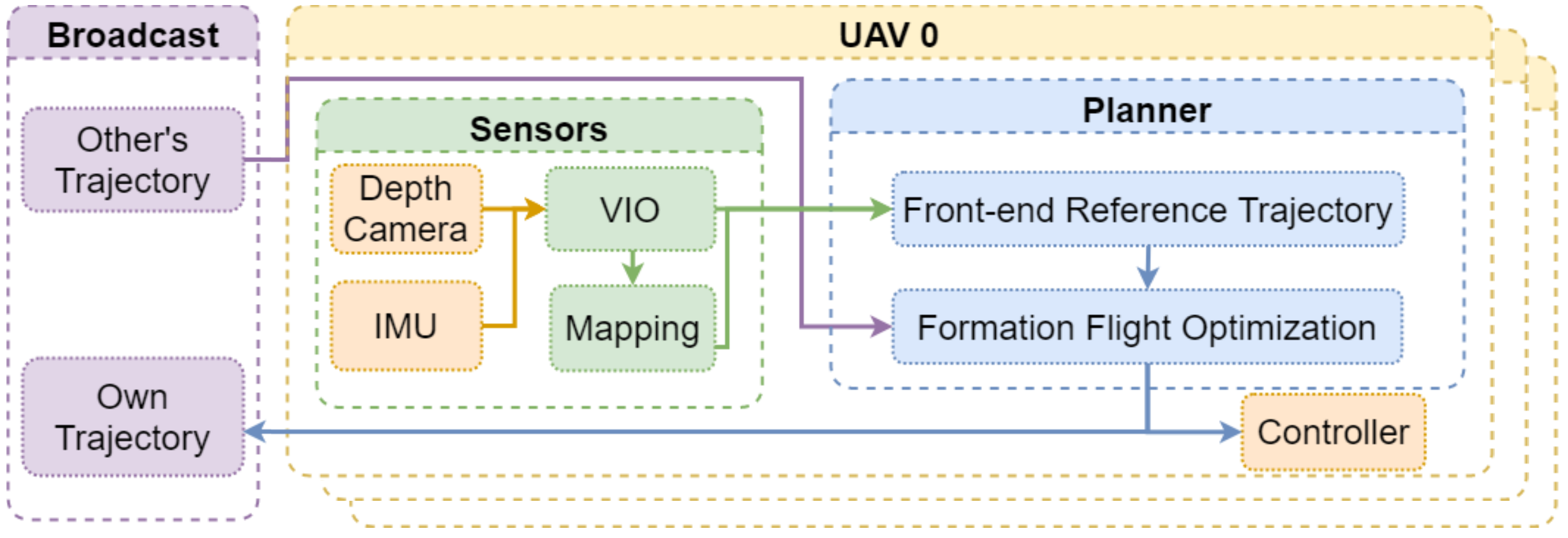}
    \end{center}
    \caption{System architecture of our distributed aerial swarm.}
    \label{fig:system}
    \vspace{-0.4cm}
\end{figure}

Our method is integrated with an autonomous distributed aerial swarm system as shown in Fig.\ref{fig:system}. The swarm shares trajectories through a broadcast network, which is the only connection among all the quadrotors. 

Each quadrotor is equipped with an Intel RealSense D435\footnote{https://www.intelrealsense.com/depth-camera-d435/} stereo camera for imagery and depth sensing.
In addition, software modules including state estimation, environment perception, trajectory planning, and flight control are all running with an onboard computer Xavier NX\footnote{https://www.nvidia.com/en-us/autonomous-machines/embedded-systems/jetson-xavier-nx/} in real-time. 

In both real-world and simulation experiments, we generate the local trajectory by solving (\ref{equ:minJ}) per second and run collision check at a frequency of 100Hz. The unconstrained optimization problem is solved by an open-source library LBFGS-Lite\footnote{https://github.com/ZJU-FAST-Lab/LBFGS-Lite}. All simulations are run on a desktop with an Intel i9-9900K CPU in real-time.

\subsection{Real-world Experiments}
\label{sec:real_world}
Real-world experiments are designed to validate the feasibility and robustness of our method. In the first experiment, as shown in Fig.\ref{fig:main}, a 2-D hexagon formation consisting of seven quadrotors successfully traverses a obstacle-rich area without any collision. Twelve cylinder obstacles with a diameter of 0.3m are placed in the area. This test demonstrates that our method is able to maintain the formation for large-scale swarms in unknown complex environments.

In the second experiment, as shown in Fig.\ref{fig:gap}, four quadrotors in a 3-D regular tetrahedron formation manage to pass through a narrow corridor safely.
During the flight, the swarm adaptively rotates and compresses the formation shape in responding to the environmental changes. This test proves that the scaling and rotational invariance provides more flexibility for formation flights in constrained spaces.  

\subsection{Simulation Experiment}
In order to testify the effectiveness of our method with large-scale irregular formations, we design a heart-shaped formation consisting of ten quadrotors. A cluttered area of $80\times20m$ with 300 cylinder obstacles and 80 circular obstacles is set up in simulation. As depicted in Fig.\ref{fig:heart}, the swarm successfully avoids the obstacles and the desired formation is well preserved during the flight.

\section{Conclusions and Future Work}
\label{sec:conclusion}
In this paper, we present a distributed swarm trajectory optimization method for formation flight in dense environments. A novel metric is proposed to measure the formation similarity, which is incorporated with a spatial-temporal optimization framework to generate swarm trajectories. The solid performance of our method in simulations and real-world experiments validates its practicality and efficiency.

In the future, we will be committed to further improving the robustness of our method. The challenges arise when the communication ranges of some robots are unpredictably narrowed. Also, in highly constrained environments, task reassignment among the robots could be necessary to resolve deadlocks timely. Finally, we hope to provide a complete solution of formation navigation in dense environments for the robotics community.

\bibliography{ICRA2022}

\begin{thebibliography}{10}
\providecommand{\url}[1]{#1}
\csname url@rmstyle\endcsname
\providecommand{\newblock}{\relax}
\providecommand{\bibinfo}[2]{#2}
\providecommand\BIBentrySTDinterwordspacing{\spaceskip=0pt\relax}
\providecommand\BIBentryALTinterwordstretchfactor{4}
\providecommand\BIBentryALTinterwordspacing{\spaceskip=\fontdimen2\font plus
\BIBentryALTinterwordstretchfactor\fontdimen3\font minus
  \fontdimen4\font\relax}
\providecommand\BIBforeignlanguage[2]{{%
\expandafter\ifx\csname l@#1\endcsname\relax
\typeout{** WARNING: IEEEtran.bst: No hyphenation pattern has been}%
\typeout{** loaded for the language `#1'. Using the pattern for}%
\typeout{** the default language instead.}%
\else
\language=\csname l@#1\endcsname
\fi
#2}}

\bibitem{marconi2012sherpa}
L.~Marconi, C.~Melchiorri, M.~Beetz, D.~Pangercic, R.~Siegwart, S.~Leutenegger,
  R.~Carloni, S.~Stramigioli, H.~Bruyninckx, P.~Doherty, A.~Kleiner,
  V.~Lippiello, A.~Finzi, B.~Siciliano, A.~Sala, and N.~Tomatis, ``The sherpa
  project: Smart collaboration between humans and ground-aerial robots for
  improving rescuing activities in alpine environments,'' in \emph{2012 IEEE
  International Symposium on Safety, Security, and Rescue Robotics (SSRR)},
  2012, pp. 1--4.

\bibitem{mahdoui2020communicating}
N.~Mahdoui, V.~Fr{\'e}mont, and E.~Natalizio, ``Communicating multi-uav system
  for cooperative slam-based exploration,'' \emph{Journal of Intelligent \&
  Robotic Systems}, vol.~98, no.~2, pp. 325--343, 2020.

\bibitem{dorling2016vehicle}
K.~Dorling, J.~Heinrichs, G.~G. Messier, and S.~Magierowski, ``Vehicle routing
  problems for drone delivery,'' \emph{IEEE Transactions on Systems, Man, and
  Cybernetics: Systems}, vol.~47, no.~1, pp. 70--85, 2016.

\bibitem{jahn2017distributed}
A.~Jahn, R.~J. Alitappeh, D.~Salda{\~n}a, L.~C. Pimenta, A.~G. Santos, and
  M.~F. Campos, ``Distributed multi-robot coordination for dynamic perimeter
  surveillance in uncertain environments,'' in \emph{2017 IEEE International
  Conference on Robotics and Automation (ICRA)}.\hskip 1em plus 0.5em minus
  0.4em\relax IEEE, 2017, pp. 273--278.

\bibitem{oh2015survey}
K.-K. Oh, M.-C. Park, and H.-S. Ahn, ``A survey of multi-agent formation
  control,'' \emph{Automatica}, vol.~53, pp. 424--440, 2015.

\bibitem{van2011RR}
J.~Van Den~Berg, S.~J. Guy, M.~Lin, and D.~Manocha, ``Reciprocal n-body
  collision avoidance,'' in \emph{Robotics research}.\hskip 1em plus 0.5em
  minus 0.4em\relax Springer, 2011, pp. 3--19.

\bibitem{van2011reciprocal}
J.~Van Den~Berg, J.~Snape, S.~J. Guy, and D.~Manocha, ``Reciprocal collision
  avoidance with acceleration-velocity obstacles,'' in \emph{2011 IEEE
  International Conference on Robotics and Automation}.\hskip 1em plus 0.5em
  minus 0.4em\relax IEEE, 2011, pp. 3475--3482.

\bibitem{bareiss2013reciprocal}
D.~Bareiss and J.~Van~den Berg, ``Reciprocal collision avoidance for robots
  with linear dynamics using lqr-obstacles,'' in \emph{2013 IEEE International
  Conference on Robotics and Automation}.\hskip 1em plus 0.5em minus
  0.4em\relax IEEE, 2013, pp. 3847--3853.

\bibitem{arul2020dcad}
S.~H. Arul and D.~Manocha, ``Dcad: Decentralized collision avoidance with
  dynamics constraints for agile quadrotor swarms,'' \emph{IEEE Robotics and
  Automation Letters}, vol.~5, no.~2, pp. 1191--1198, 2020.

\bibitem{luis2019trajectory}
C.~E. Luis and A.~P. Schoellig, ``Trajectory generation for multiagent
  point-to-point transitions via distributed model predictive control,''
  \emph{IEEE Robotics and Automation Letters}, vol.~4, no.~2, pp. 375--382,
  2019.

\bibitem{park2020efficient}
J.~Park, J.~Kim, I.~Jang, and H.~J. Kim, ``Efficient multi-agent trajectory
  planning with feasibility guarantee using relative bernstein polynomial,'' in
  \emph{2020 IEEE International Conference on Robotics and Automation
  (ICRA)}.\hskip 1em plus 0.5em minus 0.4em\relax IEEE, 2020, pp. 434--440.

\bibitem{zhou2017fast}
D.~Zhou, Z.~Wang, S.~Bandyopadhyay, and M.~Schwager, ``Fast, on-line collision
  avoidance for dynamic vehicles using buffered voronoi cells,'' \emph{IEEE
  Robotics and Automation Letters}, vol.~2, no.~2, pp. 1047--1054, 2017.

\bibitem{chen2015decoupled}
Y.~Chen, M.~Cutler, and J.~P. How, ``Decoupled multiagent path planning via
  incremental sequential convex programming,'' in \emph{2015 IEEE International
  Conference on Robotics and Automation (ICRA)}.\hskip 1em plus 0.5em minus
  0.4em\relax IEEE, 2015, pp. 5954--5961.

\bibitem{baca2018model}
T.~Baca, D.~Hert, G.~Loianno, M.~Saska, and V.~Kumar, ``Model predictive
  trajectory tracking and collision avoidance for reliable outdoor deployment
  of unmanned aerial vehicles,'' in \emph{2018 IEEE/RSJ International
  Conference on Intelligent Robots and Systems (IROS)}.\hskip 1em plus 0.5em
  minus 0.4em\relax IEEE, 2018, pp. 6753--6760.

\bibitem{zhou2021decentralized}
X.~Zhou, Z.~Wang, X.~Wen, J.~Zhu, C.~Xu, and F.~Gao, ``Decentralized
  spatial-temporal trajectory planning for multicopter swarms,'' \emph{arXiv
  preprint arXiv:}, 2021.

\bibitem{lewis1997high}
M.~A. Lewis and K.-H. Tan, ``High precision formation control of mobile robots
  using virtual structures,'' \emph{Autonomous robots}, vol.~4, no.~4, pp.
  387--403, 1997.

\bibitem{de2006formation}
M.~C. De~Gennaro and A.~Jadbabaie, ``Formation control for a cooperative
  multi-agent system using decentralized navigation functions,'' in \emph{2006
  American Control Conference}.\hskip 1em plus 0.5em minus 0.4em\relax IEEE,
  2006, pp. 6--pp.

\bibitem{balch1998behavior}
T.~Balch and R.~C. Arkin, ``Behavior-based formation control for multirobot
  teams,'' \emph{IEEE transactions on robotics and automation}, vol.~14, no.~6,
  pp. 926--939, 1998.

\bibitem{lin2013leader}
Z.~Lin, W.~Ding, G.~Yan, C.~Yu, and A.~Giua, ``Leader--follower formation via
  complex laplacian,'' \emph{Automatica}, vol.~49, no.~6, pp. 1900--1906, 2013.

\bibitem{han2013local}
Z.~Han, L.~Wang, and Z.~Lin, ``Local formation control strategies with
  undetermined and determined formation scales for co-leader vehicle
  networks,'' in \emph{52nd IEEE Conference on Decision and Control}.\hskip 1em
  plus 0.5em minus 0.4em\relax IEEE, 2013, pp. 7339--7344.

\bibitem{zhao2018affine}
S.~Zhao, ``Affine formation maneuver control of multiagent systems,''
  \emph{IEEE Transactions on Automatic Control}, vol.~63, no.~12, pp.
  4140--4155, 2018.

\bibitem{alonso2016distributed}
J.~Alonso-Mora, E.~Montijano, M.~Schwager, and D.~Rus, ``Distributed
  multi-robot formation control among obstacles: A geometric and optimization
  approach with consensus,'' in \emph{2016 IEEE international conference on
  robotics and automation (ICRA)}.\hskip 1em plus 0.5em minus 0.4em\relax IEEE,
  2016, pp. 5356--5363.

\bibitem{zhou2018agile}
D.~Zhou, Z.~Wang, and M.~Schwager, ``Agile coordination and assistive collision
  avoidance for quadrotor swarms using virtual structures,'' \emph{IEEE
  Transactions on Robotics}, vol.~34, no.~4, pp. 916--923, 2018.

\bibitem{van2017distributed}
R.~Van~Parys and G.~Pipeleers, ``Distributed model predictive formation control
  with inter-vehicle collision avoidance,'' in \emph{2017 11th Asian Control
  Conference (ASCC)}.\hskip 1em plus 0.5em minus 0.4em\relax IEEE, 2017, pp.
  2399--2404.

\bibitem{NatureReport}
M.~Tantardini, F.~Ieva, L.~Tajoli, and C.~Piccardi, ``Comparing methods for
  comparing networks,'' \emph{Scientific Reports}, vol.~9, 11 2019.

\bibitem{wang2021geometrically}
Z.~Wang, X.~Zhou, C.~Xu, and F.~Gao, ``Geometrically constrained trajectory
  optimization for multicopters,'' \emph{arXiv preprint arXiv:}, 2021.

\bibitem{zhou2019robust}
B.~Zhou, F.~Gao, L.~Wang, C.~Liu, and S.~Shen, ``Robust and efficient quadrotor
  trajectory generation for fast autonomous flight,'' \emph{IEEE Robotics and
  Automation Letters}, vol.~4, no.~4, pp. 3529--3536, 2019.

\bibitem{Press2007numerical}
W.~H. Press, S.~A. Teukolsky, W.~T. Vetterling, and B.~P. Flannery,
  \emph{Numerical Recipes 3rd Edition: The Art of Scientific Computing},
  3rd~ed.\hskip 1em plus 0.5em minus 0.4em\relax USA: Cambridge University
  Press, 2007.

\bibitem{turpin2012trajectory}
M.~Turpin, N.~Michael, and V.~Kumar, ``Trajectory design and control for
  aggressive formation flight with quadrotors,'' \emph{Autonomous Robots},
  vol.~33, no.~1, pp. 143--156, 2012.

\bibitem{parker2018pipeline}
P.~C. Lusk, X.~Cai, S.~Wadhwania, A.~Paris, K.~Fathian, and J.~P. How, ``A
  distributed pipeline for scalable, deconflicted formation flying,''
  \emph{IEEE Robotics and Automation Letters}, vol.~5, no.~4, pp. 5213--5220,
  2020.

\end{thebibliography}
\end{document}